\documentclass[12pt, double]{article}
\usepackage{times}
\usepackage{wrapfig}
\usepackage{verbatim}
\usepackage{amsmath}
\usepackage{graphicx}
\usepackage{subcaption}
\usepackage{srcltx}
\usepackage{color}
\usepackage{lineno}
\usepackage{dirtytalk}
\usepackage{amsthm}
\usepackage{authblk}
\usepackage{listings}
\usepackage{tikz}
\usepackage{epsfig,amsfonts,amssymb}
\usepackage{adjustbox}
\usepackage{multirow}
\usepackage{setspace}
\usepackage{hyperref}
\usepackage{algorithm}
\usepackage{algpseudocode}
\usepackage{amsmath}
\newcommand{\review}[1]{\textbf{#1}}

\begin{document}

\title{\Large Tighten The Lasso: A Convex Hull Volume-based Anomaly Detection Method}
 \author[1,*,x]{Uri Itai} \author[2]{Asael Bar Ilan} \author[3,4,x]{Teddy Lazebnik}  \affil[1]{Department of Mathematics, The Guangdong Technion-Israel Institute of Technology, China} \affil[2]{Department of Computer Science, Open University of Israel, Israel} \affil[3]{Department of Information Systems, University Of Haifa, Israel} \affil[4]{Department of Computing, Jonkoping University, Sweden} \affil[*]{Corresponding author: \texttt{uri.itai@gmail.com}}
 \affil[x]{These authors contributed equally.}
\date{\today}

\maketitle

\thispagestyle{empty}

\begin{abstract}
\review{Detecting out-of-distribution (OOD) data is a critical task for maintaining model reliability and robustness. In this study, we propose a novel anomaly detection algorithm that leverages the convex hull (CH) property of a dataset by exploiting the observation that OOD samples marginally increase the CH's volume compared to in-distribution samples. Thus, we establish a decision boundary between OOD and in-distribution data by iteratively computing the CH’s volume as samples are removed, stopping when such removal does not significantly alter the CH's volume. The proposed algorithm is evaluated against seven widely used anomaly detection methods across ten datasets, demonstrating performance comparable to state-of-the-art (SOTA) techniques. Furthermore, we introduce a computationally efficient criterion for identifying datasets where the proposed method outperforms existing SOTA approaches.} 
\end{abstract}

\pagestyle{myheadings}
\markboth{Draft: \today}{Draft: \today}
\setcounter{page}{1}

\section{Introduction}
\label{sec:introduction}
\review{Anomaly Detection (AD) plays a pivotal role across machine learning, data analysis, statistics, and other domains involving quantitative data \cite{r1}. It focuses on identifying patterns, observations, or events that markedly deviate from expected norms, often termed \emph{anomalies} or \emph{outliers}~\cite{chandola2009anomaly,r2,r3}. Such anomalies frequently correspond to rare yet important occurrences~\cite{pang2021deep, xu2019recent} and have diverse applications, including the early detection of mechanical faults~\cite{chandola2009anomaly}, financial fraud prevention~\cite{ahmed2016survey}, cybersecurity~\cite{ten2011anomaly}, enhanced decision-making~\cite{prasad2010anomaly}, as well as powering real-time clinical alert systems~\cite{habeeb2019real}.}

In a data-driven context, AD is commonly regarded as a subfield of machine learning (ML) and can be mathematically framed as a binary classification problem, where data is categorized as either normal or anomalous \cite{jia2019anomaly,gornitz2013toward,lazebnik2025pulling}. However, due to the inherent rarity of anomalies, conventional \say{normal} classification models often underperform in this task, necessitating the development of specialized algorithms tailored for anomaly detection \cite{nassif2021machine}.

\review{Indeed, various groups of anomaly detection (AD) algorithms have been developed, each tailored to different types of data and problem contexts \cite{fernando2021deep,himeur2021artificial,r4,r5}. These algorithms can be broadly categorized into three primary types: statistical-based methods, machine learning (ML)-based methods, and distance-based methods. These models focus on the hyperplane which is a flat \(d-1\)-dimensional surface in a \(d\)-dimensional space, separating normal from abnormal data by positioning a single linear boundary. Its advantages are low computational complexity, simplicity, and good performance for linearly separable data, but it cannot directly capture complex or curved data shapes without kernel transformations. In contrast, a hypervolume—often represented by the convex hull—forms the smallest convex region enclosing all normal data points, allowing it to model arbitrary convex shapes in the feature space. This provides higher flexibility and the ability to tightly wrap around the data distribution, which can improve detection accuracy for non-linear patterns, while being more computationally expensive.}

The traditional method of determining anomalies by taking $k$ standard deviations away from the mean is extended using the Z-score (T statistics), in which the confidence interval is represented as an ellipsoid. This advancement paves the way for more refined, statistically-based techniques, including Gaussian mixture models \cite{li2016anomaly}, Mahalanobis distance \cite{barnard2000detecting}, and hypothesis testing \cite{cohen2015active}.
This method concerning the convex hull was done in \cite{costa2013partially}.
These methods rely on the presumption that the data adheres to a specific statistical distribution, thereby identifying any points that significantly diverge from this distribution as anomalies.

ML-based methods can be classified into supervised, semi-supervised, and unsupervised approaches \cite{alpaydin2020introduction}. Supervised methods, such as neural networks or support vector machines, require labeled datasets containing normal and abnormal instances. However, obtaining these labels is often impractical in real-world scenarios due to the high cost and difficulty associated with labeling data \cite{wang2014new}. In contrast, semi-supervised methods, such as  One-class SVM \cite{manevitz2001one}, One-class deep neural networks \cite{oza2018one}, and auto-encoders \cite{zhou2017anomaly}, as well as local outlier factor \cite{cheng2019outlier}, are trained exclusively on normal data and detect anomalies by identifying instances that do not conform to the learned patterns \cite{boukerche2020outlier}. Additional methods include contracting a gain function \cite{novello2024improving} or constructing a distribution \cite{costa2013partially}.

Another approach involves the use of Voronoi diagrams \cite{blaise2022group}. In this method, the data is partitioned into bins, and points that do not fit within these bins are flagged as anomalies. The computation of each cell in the diagram is based on the convex hull.

Distance-based methods, such as K-nearest neighbors (KNN) \cite{peterson2009k} and clustering techniques like DBSCAN \cite{schubert2017dbscan}, involve measuring the distance or similarity between data points. Data points far from clusters or with few neighboring points are considered anomalies. These methods are typically limited in scalability, as they struggle to handle large datasets due to their computational complexity \cite{mandhare2017comparative}. Moreover, distance-based methods tend to perform poorly in non-uniform data distributions, as they assume anomalies are far from most data points. 

An alternative approach to AD utilizes the concept of a convex hull, defined as the smallest convex set that contains a given set of points in a metric space \cite{avis1995how}. Simply put, a convex hull is a set of outermost nails on a board stretching a rubber band around a set of nails when released. Computing the convex hull plays an important role in many fields of computer science, such as computer graphics \cite{barber1996quickhull} and computer-aided design \cite{barnhill2014computer}. This geometric framework leads to convex hull-based algorithms for AD, which offer an efficient way to identify anomalies by leveraging geometric properties \cite{wang2022anomaly}. A convex hull, representing the boundary of the normal data distribution, delineates the smallest convex shape enclosing a set of points in multi-dimensional space \cite{wang2022anomaly}. Any data point outside this convex boundary is considered anomalous \cite{costa2013partially}. These algorithms are particularly powerful because they provide a clear, interpretable geometric definition of normal behavior, making it easy to visualize and understand the separation between normal and anomalous data. Moreover, convex hull-based methods are adaptable to high-dimensional and complex datasets as they require only a well-defined distance metric, which is relatively straightforward in most real-world applications \cite{li2016improving,olteanu2023meta}. Several studies already used convex-hull-based AD algorithms, showing promising results \cite{ch_g_1,ch_g_2,ch_g_3}. For instance, Liu at al. \cite{liu2009novel} proposed adding a discount factor to the convex hull to avoid overfitting on the training dataset. Casale et al. \cite{casale2014approximate} reduce the computation time of the convex hull as an AD algorithm using the approximate polytope ensemble technique. In an applied context, He et al. \cite{he2020kernel} propose a novel classifier, the Kernel Flexible and Displaceable Convex Hull-based Tensor Machine, designed for gearbox fault diagnosis using multi-source signals. This classifier uses a convex hull approach in tensor feature space to classify feature tensors, demonstrating improved robustness and effectiveness in identifying gearbox faults with small sample sizes.

Although efficient, existing algorithms lack a definitive mechanism for determining the optimal convex hull shape for a given dataset without introducing additional assumptions. Furthermore, their practical applications remain ambiguous within the broader context of state-of-the-art anomaly detection (AD) algorithms.

In this study, we propose a novel parameter-free AD algorithm grounded in the convex hull family of methods. This algorithm balances two configurations: the convex hull encompassing the entire dataset and the convex hull computed for a subset of the data, excluding the most "distant" samples. By leveraging this approach, we introduce a topology-agnostic AD algorithm that maintains computational efficiency and resource requirements comparable to other state-of-the-art AD algorithms while exhibiting robustness to complex data geometries.

The proposed method not only identifies anomalies but also quantifies the extent to which a given instance deviates from the norm. This metric is instrumental in distinguishing between tail events and out-of-distribution instances, a distinction of particular importance in time series analysis \cite{blazquez2021review} and online data processing \cite{subramaniam2006online}. While a detailed examination of this distinction lies beyond the scope of the present paper, it will be addressed in future research.

We conducted extensive experiments using ten real-world datasets spanning various domains to evaluate the proposed method. We benchmarked the performance of our model against seven state-of-the-art AD algorithms representing all three principal categories of AD methods. The results demonstrate. The Convex Hull algorithm demonstrates strong performance, surpassing all competing methods except for Isolation Forest and Local Outlier Factor. However, Isolation Forest does not account for distances and angles and remains invariant under shearing transformations, which may not always be desirable. In contrast, the Local Outlier Factor is a local algorithm, meaning it primarily identifies local anomalies rather than global ones. As a result, it may fail to detect global anomalies. Consequently, the Convex Hull-based anomaly detection algorithm proves to be effective in scenarios that require a global geometric perspective.

The rest of the paper is organized as follows. Section \ref{sec:related_work} reviews the state-of-the-art AD algorithms as well as the convex hull family of algorithms and provides examples of geometry-based AD algorithms for one and two dimensions. Section \ref{sec:algorithm} formally introduces the proposed convex-hull-based AD algorithm with an analytical analysis of the algorithm. Section \ref{sec:experiments} outlines the experiments on real-world data comparing the proposed algorithms with current state-of-the-art AD algorithms and also the sensitivity analysis of the proposed algorithm. Finally, Section \ref{sec:conclusions} discusses the applicative outcomes of this study and suggests possible future research. 

\section{Related Work}
\label{sec:related_work}

This section presents a comprehensive review of the definition of the convex hull, followed by an exploration of the algorithms commonly utilized for its numerical computation. Subsequently, an in-depth examination of the current state-of-the-art anomaly detection (AD) algorithms is provided, focusing on their methodologies, applications, and performance characteristics.

\subsection{Convex hull}
\label{sec:rw_convex_hull}

The convex hull holds significant importance across various research domains, including computer graphics \cite{jayaram2016convex}, computational geometry \cite{lee1984computational}, optimization \cite{lachand2005minimizing}, and numerous other fields.

The convex hull is the smallest convex set that completely encloses a given set of points, minimizing the hyper-volume enclosed by the set. Formally, for a set of points \( \{\boldsymbol{x}_i\}_{i=1}^n \subset \mathbb{R}^d \), the convex hull is the minimal convex polytope containing all points, and it is expressed as \cite{cristescu2013non}:

\begin{equation}
CH(S) = \left\{ \sum_{i=1}^{n} \alpha_i \boldsymbol{x}_i \mid \alpha_i \geq 0 \, \wedge \, \sum_{i=1}^{n} \alpha_i = 1 \right\},
\end{equation}

where \( \alpha_i \) represents the non-negative scalar coefficients associated with each point \( \boldsymbol{x}_i \) in the set \( S \).

\subsubsection{Computing convex hull}
Various methods have been proposed to compute the convex hull of an arbitrary set of points, leveraging the convex hull's inherent properties of convexity. Among these, three widely utilized algorithms are noteworthy: the Gift Wrap algorithm \cite{chan1996optimal}, the Graham Scan \cite{xu2010concave}, and the Quickhull algorithm \cite{barber1996quickhull}.

The Gift Wrap algorithm \cite{chan1996optimal}, also known as Jarvis March, constructs the convex hull incrementally by identifying boundary points one at a time. Starting with the leftmost point, which is guaranteed to be part of the convex hull, the algorithm iteratively selects the point that forms the smallest counterclockwise angle with the line segment formed by the current hull boundary. This selection ensures that the boundary progresses in a \say{wrapping} motion, ultimately enclosing all points. The hull is complete once the algorithm loops back to the starting point. The simplicity of the Gift Wrap algorithm makes it a natural choice for small datasets or cases where the number of points on the hull is small. However, its time complexity, \(\mathcal{O}(nh)\), where \(n\) is the total number of points and \(h\) is the number of points on the hull, can become prohibitive for larger datasets with dense point distributions.

Graham’s Scan algorithm \cite{xu2010concave} begins by identifying a reference point, typically the point with the smallest \(y\)-coordinate (or the leftmost point in case of ties). The algorithm then sorts all points by their polar angle relative to this reference point, ensuring a natural order for constructing the hull. Using a stack, Graham’s Scan processes these points sequentially, adding points to the stack while ensuring that each addition maintains the convexity of the boundary. If a new point causes the boundary to form a concave angle, points are removed from the stack until convexity is restored. The reliance on sorting gives Graham’s Scan a time complexity of \(\mathcal{O}(n \log n)\), making it particularly well-suited for static datasets where sorting overhead can be amortized.

The Quickhull algorithm employs a divide-and-conquer approach to compute the convex hull of a given set of points. The process begins by identifying the points with minimum and maximum \(x\)-coordinates defining the initial boundary segment. The algorithm then determines the point farthest from this segment, effectively partitioning the dataset into two subsets: points located to the left and right of the segment. This procedure is applied recursively to each subset, iteratively identifying the points that constitute the convex hull. Although Quickhull is well-suited for datasets with non-uniform spatial distributions, its worst-case computational complexity is \(\mathcal{O}(n^2)\), which occurs when the point distribution leads to excessive recursion.

These algorithms form the basis for efficient convex hull computation and have been extended to handle dynamic datasets. For instance, updating the convex hull after removing a single point can often be achieved in \(\mathcal{O}(n + k)\), where \(k\) is the number of points on the hull boundary, avoiding the need for a complete recomputation. Their adaptability and efficiency ensure these methods remain central to computational geometry and its applications in higher dimensions.

Chan's algorithm \cite{chan1996optimal} computes the convex hull with a time complexity of \( O(n \log n + n^{\lfloor d/2 \rfloor}) \), making it computationally efficient for large datasets. An alternative method is described by Nielsen and Nock \cite{nielsen2009sided}, which integrates the efficiency of divide-and-conquer strategies with the precision of geometric methods. This ensures that the convex hull can be computed in optimal time for various applications.

\subsubsection{Convex hull for anomaly detection}
Identifying anomalies using the convex hull is based on the observation that anomalous points tend to expand the convex hull's boundaries. From an energy perspective, this approach can be interpreted as a trade-off analysis, wherein the inclusion of additional data points results in a transition from a set with a minimal convex hull to an expanded configuration encompassing these new points. This shift highlights the balance between maintaining a compact representation and accommodating potential anomalies within the dataset.
Using a support vector machine for constructing boundaries hyperplane was done by Zhang and Gu~\cite{zhang2007ch} and Wand et al.  \cite{wang2022anomaly}. Using the convex hull to improve the calculation of the mean and variance to detect anomalies was done by Costa et al. \cite{costa2013partially}. Blaise et al. \cite{blaise2022group} suggest a method to detect a group of anomalies using the Voroi diagram and the convex hull. 

\subsection{Anomaly Detection}
\label{sec:rw_anomaly_detection}
Anomaly detection (AD) plays a critical role in database projects. However, due to the predominantly unsupervised nature of AD, there is often no definitive solution to AD tasks in many real-world applications \cite{ghahramani2003unsupervised}. \review{an be formally defined as the task of identifying data instances 
\( x \in \mathcal{X} \) whose characteristics deviate significantly from those of the majority of the data, according to a predefined or learned notion of normality. AD is often defined as follows. Given a dataset  \( \mathcal{D} = \{x_1, x_2, \ldots, x_n\} \) sampled from an unknown distribution \( P(X) \).  The goal is to learn a scoring function \(s: \mathcal{X} \rightarrow \mathbb{R}\) such that higher values of \( s(x) \) indicate a higher likelihood of \( x \) being an anomaly. Notably, this definition is not computationally strict and reveals the adoptive nature of the definition in different applied settings.} As a result, a variety of methods have been developed over the years, each employing distinct strategies to identify anomalies based on specific assumptions regarding the properties of the data, the anomalies, or both. Among these methods, several algorithms have gained widespread adoption:

Isolation Forest \cite{cheng2019outlier} isolates individual data points through recursive partitioning of the data space, identifying anomalies based on the speed at which they can be separated. It constructs an ensemble of isolation trees, where data points that are isolated with shorter average path lengths are flagged as anomalies. \textit{Isolation Forest} is particularly effective when anomalies are sparse and well-separated from normal data points, especially in high-dimensional spaces. However, its performance may be compromised when anomalies are densely clustered or exhibit intricate patterns, as random splits may fail to capture these underlying structures. Moreover, this method relies solely on the ordinal ranking of each variable, disregarding distances and inter-variable relationships. As a result, its effectiveness may be diminished in situations where these factors are critical. The method operates based on the topology of the data, meaning that geometrical transformations, such as stretching or shrinking, do not affect the results. Nonetheless, in many practical applications, the geometry of the data plays an important role, and omitting this consideration may lead to false discoveries of abnormalities.

\textit{Single-Class SVM} \cite{shin2005one} detect anomalies by learning a boundary around normal data. Trained on data from a single class, these methods assume points outside the learned boundary are anomalous. \textit{Single-Class SVMs} construct a hyperplane that maximally separates normal data from the origin.

Gaussian Mixture Models (GMM) \cite{li2016anomaly} detect anomalies by modeling data as a mixture of Gaussian distributions, identifying points with low likelihood under the model as anomalies. \textit{GMM} is suitable for data clusters that approximate Gaussian shapes but may struggle with non-Gaussian clusters or complex distributions. This can be a generalization of the Z-score. Instead of a single bell in the Gaussian distribution, there exist a few. In real-life data, this is not common. 

Local Outlier Factor (LOF) \cite{alghushairy2020review} measures the local density deviation of a data point relative to its neighbors, identifying anomalies in areas of significantly lower density. \textit{LOF} is useful for data with local clusters or varying densities, but may struggle with uniform-density data or when clear neighborhood structures are absent. This method detects anomalies without considering the global structure of the data. Nonetheless, detecting abnormalities globally is very important in many cases. 

Density-Based Spatial Clustering of Applications with Noise (DBSCAN) \cite{schubert2017dbscan} groups data into dense regions, identifying points in sparse regions as anomalies. \textit{DBSCAN} is effective for datasets with varying densities and distinct clusters but may perform poorly on data with uniform density or overlapping clusters.

K-means \cite{munz2007traffic} partitions data into a set number of clusters, flagging points with high distances from the nearest cluster center as anomalies. This method assumes spherical clusters and works best when clusters are compact and anomalies are far from cluster centers but are limited by varying cluster shapes or densities.

Mean Shift \cite{yang2021mean} iteratively shifts data points toward high-density regions, identifying clusters as density peaks and labeling points outside these clusters as anomalies. \textit{Mean Shift} is effective when data has distinct density peaks but may be less effective in uniformly distributed data or data lacking prominent clusters.

The primary limitation of the clustering process is its inherent instability, which is highly sensitive to the initial conditions.

\section{Convex Hull Method for Anomaly Detection}
\label{sec:algorithm}
\begin{figure}
    \centering
    \includegraphics[width=0.99\linewidth]{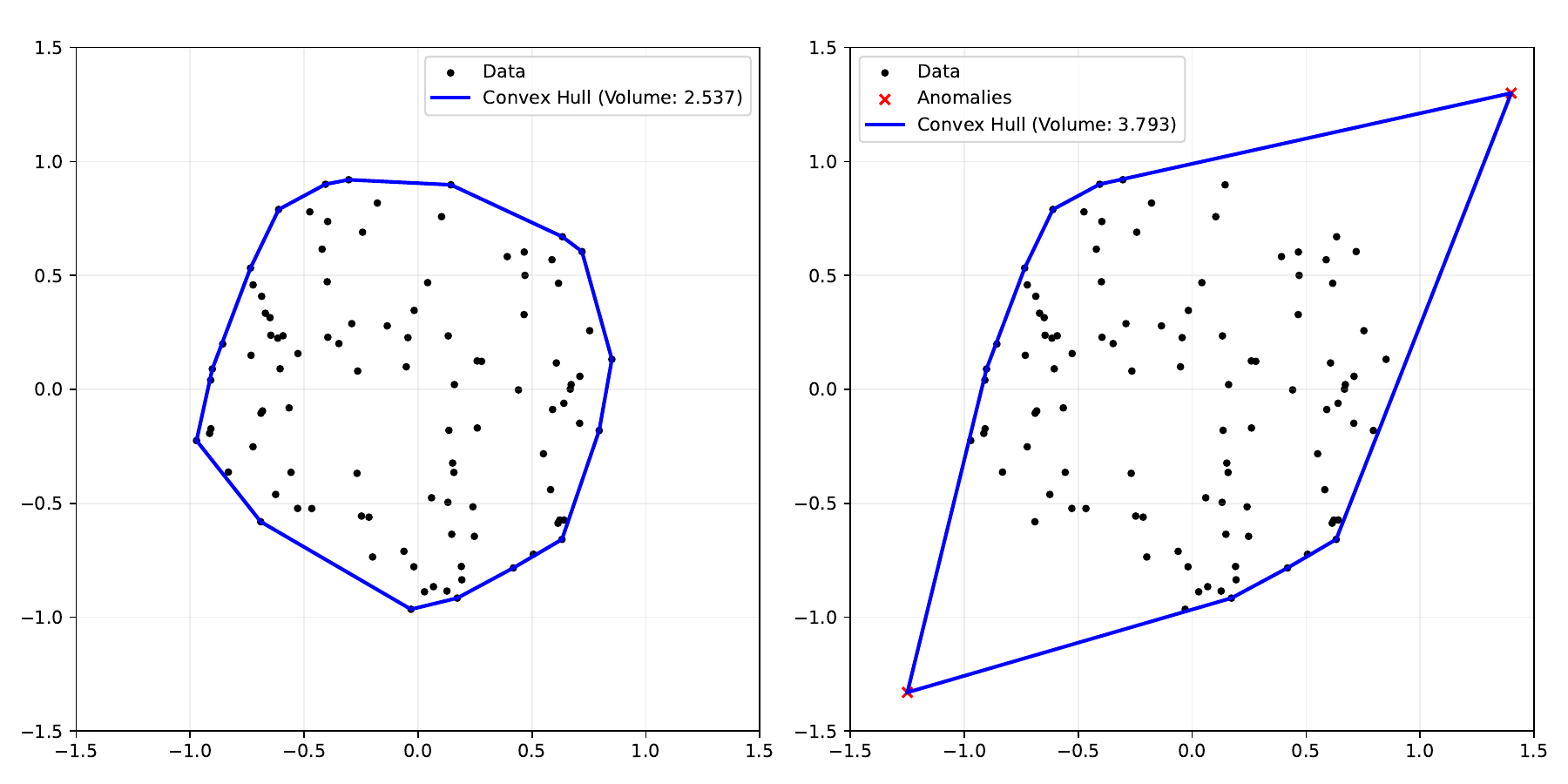}
    \caption{A simplistic example of a 2D dataset with an anomaly and its influence on the CH's volume. On the left, there is the dataset without anomalies with a volume of 2.537 while, as shown on the right, two anomaly samples increase the CH's volume to 3.793, almost 50\% increase.}
    \label{fig:example}
\end{figure}

A volume-based convex hull method provides a robust framework for anomaly detection by leveraging the geometric properties of the data distribution to identify points that deviate significantly from the majority. The convex hull's volume measures the dataset's spatial extent within \( n \)-dimensional space. Anomalies are often located at the periphery of the data distribution, where their separation from the dense data core disproportionately inflates the convex hull's volume. \review{Fig. \ref{fig:example} presents a simplistic example of a 2D dataset with an anomaly and its influence on the CH's volume. }

The proposed method identifies compact and densely distributed subsets of data by minimizing the convex hull volume while maximizing the number of enclosed points. This approach effectively excludes anomalous points. However, balancing these conflicting objectives—maximizing data inclusion while minimizing the convex hull volume—poses significant challenges, akin to the sorites paradox \cite{hyde2011sorites} when considered in extreme scenarios. 

To address these challenges, a well-defined stopping condition must be established for the iterative removal or addition of points. Alternatively, appropriate weights can be assigned to these competing objectives within an optimization framework to achieve a balance between the two goals.

\review{Formally, to define an algorithm based on the above motivation, we first define \(S_p\) as the subset of \(S\), a set \(S \subseteq \mathbb{R}^n\), that maximizes the volume-based objective. The problem can be stated as the following combinatorial optimization task:}
\begin{equation}\label{Eq:opt_problem}
\max_{S_p \subseteq S} \; f(S_p) = |S_p| - \lambda \, \mathrm{vol}\big(CH(S_p)\big), \quad \lambda > 0,
\end{equation}
\review{where \(CH(S_p)\) represents the convex hull of the subset \(S_p\), and \(\mathrm{vol}(CH(S_p))\) denotes its volume. The objective in Eq. (\ref{Eq:opt_problem}) quantifies the trade-off between the size of the selected subset \(|S_p|\) and the compactness of its convex hull, with \(\lambda\) acting as a sensitivity parameter. Larger \(\lambda\) values prioritize minimizing the convex hull volume, aiding anomaly isolation, while smaller \(\lambda\) values tolerate broader variations in the data distribution.}

Algorithm \ref{Alg:Ab_CH_Mod} shows a pseudo-code of the proposed algorithm which accepts a dataset (\(S\)), stopping criteria (\(SC\)), and an algorithm to compute the convex hull \(CH\); and returns the subset of samples in the dataset which are not anomaly-free. \review{At each iteration, the convex hull \(CH(S_p)\) is fully recomputed from the current subset \(S_p\) after the removal of each candidate point \(p\). No incremental hull update strategy is used; the recomputation is performed in full to ensure correctness and to maintain consistency with the volume evaluation.}

\begin{algorithm}[H]
\caption{Modified Volume-Based Convex Hull Anomaly Detection}\label{Alg:Ab_CH_Mod}
\begin{algorithmic}[1]
\Require Dataset \( S \subseteq \mathbb{R}^n \), stopping criteria \( SC \), convex hull method \( \text{CH} \)
\State Initialize \( S_p \gets S, r \gets \emptyset \)
    \State \( S_h \gets \text{CH}(S_p) \)
    \State \( vol_c \gets volume(S_h) \)
    \State \( vol_m \gets vol_c \)
    \State \( S_h^{new} \gets S_h \)
    \State \( r_n \gets \emptyset \)
\Repeat
    \For{\( p \in S_h \)}
        \State \( S_p^n \gets S_p \setminus \{p\} \)
        \State \( S_h^n \gets \text{CH}(S_p^n) \)
        \State \( vol_n \gets volume(S_h^n) \)
        \If{\( vol_n \leq vol_m\)}
            \State \( S_h^{new} \gets S_h^n \)
            \State \( vol_m \gets vol_n \)
            \State \( r_n \gets p \)
        \EndIf
    \EndFor
    \State \( S_h \gets S_h^{new} \) 
    \State \( r \gets r \cup r_n \) 
    \If{\( SC \vee |r| = |S|\)}
        \State \Return \( S \setminus r \)
    \EndIf
\Until{True}
\end{algorithmic}
\end{algorithm}

Practically, one can use multiple stopping conditions, including the Elbow Point Method \cite{thorndike1953belongs}, which locates the inflection point in a cost function curve, and the Akaike Information Criterion (AIC) 
\cite{akaike2011akaike}. \review{The Elbow Point ethod is a heuristic technique for determining the optimal number of components (e.g., clusters) in a model by identifying the inflection point in a cost function curve, such as the within-cluster sum of squares. The method relies on plotting the cost function against the number of components and selecting the point beyond which additional components yield diminishing returns in performance improvement. This point, resembling an \say{elbow} in the curve, represents a balance between model complexity and fit. The Akaike Information Criterion (AIC) \cite{akaike2011akaike} is a statistical measure for model selection that balances goodness-of-fit and model complexity. It is defined as \(\mathrm{AIC} = 2k - 2\ln(\hat{L})\), where \(k\) is the number of estimated parameters in the model and \(\hat{L}\) is the maximum likelihood value. Lower AIC values indicate models that better trade off explanatory power with parsimony, making the criterion particularly useful for comparing models fitted to the same dataset.} 

The computational complexity of the proposed algorithm is determined by the iterative nature of its optimization process and the operations performed within each iteration. The algorithm comprises an outer loop that iterates until the stopping condition \( SC \) is satisfied or all points are removed from the dataset (\( |r| = |S| \)). Let \( T \) denote the total number of iterations. 

Within each iteration, the algorithm computes the convex hull \( S_h \) of the remaining dataset \( S_p \) and evaluates all points within the current convex hull. For a dataset of size \( n \), computing the convex hull requires \( O(|S_p|^2) \) operations in the worst case, where \( |S_p| \leq n \). The algorithm evaluates each point \( p \) in the convex hull \( S_h \) by temporarily removing it to calculate the volume of the resulting convex hull \( S_h^n \). Each convex hull computation for \( S_p^n = S_p \setminus \{p\} \) also has a complexity of \( O(|S_p|^2) \). 

Given that \( |S_h| \), the number of points in the convex hull, can be as large as \( n \), evaluating all points in \( S_h \) requires \( O(|S_h| \cdot |S_p|^ 2) \) operations per iteration. In the worst-case scenario where \( |S_h| = n \) and \( |S_p| = n \), this results in \( O(n \cdot n^2) = O(n^3) \) operations for one iteration. The total number of iterations \( T \) is bounded by \( n \) in the worst case, as at least one point is removed from \( S_p \) in each iteration. Thus, the overall time complexity of the algorithm becomes \( O(T \cdot n^3) = O(n^4) \).

The memory requirements of the algorithm are primarily determined by the need to store the input dataset, the convex hull \( S_h \), and intermediate results. The input dataset \( S \subseteq \mathbb{R}^d \), consisting of \( n \) points in \( d \)-dimensional space, requires \( O(n \cdot d) \) memory. 

The convex hull \( S_h \) requires storage proportional to the number of vertices in the hull. In the worst-case scenario, where all \( n \) points are part of \( S_h \), this also necessitates \( O(n \cdot d) \) memory. Temporary storage is required for subsets such as \( S_p^n = S_p \setminus \{p\} \), which similarly demands \( O(n \cdot d) \) memory. 

The computation of the volume introduces a negligible constant memory overhead. Therefore, the memory complexity is primarily driven by the storage requirements for the dataset and the convex hull, resulting in an overall memory complexity of \( O(n \cdot d) \).

\section{Experiments}
\label{sec:experiments}
\review{In this section, we present the experiments conducted to evaluate the proposed CH anomaly detection algorithm. All experiments were executed on a dedicated server running Windows 10 Pro (64-bit), equipped with an Intel Core i7-10700 CPU (8 cores, 16 threads, 2.90 GHz base frequency), 32 GB of DDR4 RAM, and a 1 TB NVMe SSD for storage.}

\subsection{Datasets}
Table \ref{table:datasets} presents the datasets used in this study with their number of rows, cols, and portion of tagged anomalies. The datasets range from small ones with as few as 148 rows and up to medium ones with over 60 thousand rows. The number of columns also ranges from seven to forty-one columns, representing a relatively wide range of configurations. In addition, we also present the datasets that are considered appropriate to the CH algorithms, denoted by \textit{Yes} in the last column, as these show a large reduction in the CH's volume over the first 10 steps of the algorithm. 

\begin{table}[h!]
\centering
\begin{tabular}{p{0.16\textwidth}p{0.24\textwidth}p{0.1\textwidth}p{0.1\textwidth}p{0.1\textwidth}p{0.1\textwidth}p{0.1\textwidth}}
\hline \hline
\textbf{Name} & \textbf{Description} & \textbf{\# Samples} & \textbf{\# Cols} & \textbf{Portion of Anomalies} & \textbf{CH-friendly} \\ \hline\hline
Glass  & Glass identification data & 214 & 7 & 4.2\% & No \\
Ionosphere  & Radar data for ionosphere detection & 351 & 32 & 35.9\% & No\\
Lymphography  & Medical lymphography data & 148 & 19 & 4.1\% & No\\
PenDigits  & Handwritten digit recognition data & 9,868 & 16 & 0.2\%& No \\
Shuttle  & NASA shuttle anomaly data & 1,013 & 9 & 1.3\% & No \\
WBC  & Wisconsin breast cancer data & 454 & 9 & 2.2\%& No \\
Waveform  & Synthetic waveform data & 3,443 & 21 & 2.9\%  & Yes \\
KDDCup99  & Network intrusion detection dataset & 60,632 & 41 & 0.4\%& Yes \\
WDBC  & Wisconsin diagnostic breast cancer & 367 & 30 & 2.7\% & Yes \\
WPBC  & Wisconsin prognostic breast cancer & 198 & 33 & 23.7\% & Yes \\ \hline \hline
\end{tabular}
\caption{An overview of the datasets used in this study.}
\label{table:datasets}
\end{table}

\subsection{Convex hull anomaly detection implementation}
In order to explore different implementations of the proposed CH anomaly detection algorithm, we consider two properties: a dimension reduction algorithm and a stop condition. For the dimension reduction, we adopted the popular principal component analysis (PCA) \cite{roweis1997algorithms} and the t-distributed stochastic neighbor embedding (t-SNE) \cite{wattenberg2016use} algorithms. As a default, we assume both reduce to a two-dimensional space. We used this step to obtain a feasible computational time for the proposed algorithm. For the stop condition, we adopted three strategies: Naive, Elbow, and Optimal. For the Naive strategy, we stopped the process once the change in the CH's volume was less than one percent of the original change (i.e., the change in the CH's volume between the original dataset and after removing the first data point). For the Elbow strategy, we performed the entire CH's computation, up to three data points, and took the elbow point of the CH's volume profile \cite{bholowalia2014ebk}. Finally, the Optimal strategy is an unrealistic one and is used only to explore the model's performance, in which the number of anomaly points is known in advance, and the CH stops after the same number of steps. \review{Importantly, we used the Euclidean distance \cite{r6} for all the experiments.}

\subsection{Performance}
Table \ref{table:p1} presents the performance of all six CH configurations (for two-dimensionality reduction methods over three stooping strategies) and seven baseline anomaly detection algorithms in terms of their accuracy, \(F_1\) score, recall, precision, \review{area under the receiver operating characteristic (ROC) curve (AUC), and computation time (CT)}. Notably, the proposed Convex Hull algorithm performs comparably to the Isolation Forest model, which is considered the current state-of-the-art and obtains the highest \(F_1\) score of 0.2619. More precisely, the Convex Hull algorithm with the optimal stopping strategy and t-SNE obtained the highest precision (and accuracy), followed by the other versions of the Convex Hull, and only then by the Isolation forest algorithm. The DBSCAN algorithm obtains an almost perfect recall of 0.9968 but overall produces extremely poor results with an \(F_1\) score of 0.0528. Expectedly, the optimal, elbow, and naive stop strategies result in a monotonic decrease in performance in terms of all metrics. In addition, the t-SNE produces better results in terms of precision and \(F_1\) score but worse in terms of recall compared to the PCA method. \review{Notably, the proposed CH algorithm is implemented in Python programming language without performance optimization, while the baseline algorithms are implemented in the \textit{C} program language with performance optimization.}

\begin{table}[h!]
    \centering
    \begin{tabular}{lcccccc}
        \hline \hline
        \textbf{Algorithm} & \textbf{Accuracy} & \textbf{F1 Score} & \textbf{Recall} & \textbf{Precision}  & \textbf{AUC}  & \textbf{CT} \\
        \hline \hline
        Convex Hull, Naive + PCA       & 0.8739 & 0.2181 & 0.2702 & 0.2112 & 0.5730 & 17.0416 \\
        Convex Hull, Naive + t-SNE     & 0.9150 & 0.2337 & 0.2244 & 0.2240 & 0.5764 & 55.5415 \\
        Convex Hull, Elbow + PCA       & 0.8739 & 0.2181 & 0.2702 & 0.2112 & 0.6016 & 8.7743 \\
        Convex Hull, Elbow + t-SNE     & 0.9163 & 0.2335 & 0.2203 & 0.2261 & 0.5792 & 69.0557 \\
        Convex Hull, Optimal + PCA     & 0.8739 & 0.2181 & 0.2702 & 0.2112 & 0.5730 & 16.0395 \\
        Convex Hull, Optimal + t-SNE   & \textbf{0.9154} & 0.2343 & 0.2229 & \textbf{0.2278} & 0.5782 & 46.4837 \\
        \hline
        Isolation Forest              & 0.9077 & \textbf{0.2619} & 0.6681 & 0.1812 & \textbf{0.8866} & \textbf{1.1620} \\
        Local Outlier Factor          & 0.9003 & 0.2110 & 0.6135 & 0.1438  & 0.8113 & 2.4859 \\
        One-Class SVM                 & 0.7665 & 0.1703 & 0.5824 & 0.1173  & 0.7191 & 7.1903 \\
        Gaussian Mixture Models       & 0.4131 & 0.0894 & 0.5203 & 0.0612  & 0.8863 & 2.2089 \\
        K-means                       & 0.3727 & 0.0736 & 0.5086 & 0.0480  & 0.4655 & 2.4578 \\
        DBSCAN                        & 0.0346 & 0.0528 & \textbf{0.9968} & 0.0283 & 0.5210 & 32.9040 \\
        Mean Shift                    & 0.0622 & 0.0497 & 0.9090 & 0.0267 & 0.4723 & 4.739 \\
        \hline \hline
    \end{tabular}
    \caption{Performance of the models with comparison to baseline algorithms. The best outcome for each metric is highlighted in bold.}
    \label{table:p1}
\end{table}

Table \ref{table:p2} presents a similar analysis to Table \ref{table:p1} but focuses only on the four datasets that are deemed Convex Hull friendly (see Table \ref{table:datasets}). In such settings, the Convex Hull with an optimal stooping strategy with the t-SNE obtains the highest \(F_1\) score and precision of 0.1698 and 0.1491, respectively. Even the more realistic configuration of the algorithm with the elbow point stooping condition outperforms all other algorithms in terms of \(F_1\) score and precision of 0.1657 and 0.1457, respectively. Like before, optimal, elbow, and naive stooping strategies result in a monotonic decrease in performance in terms of all metrics. In addition, the t-SNE produces better results in terms of precision and \(F_1\) score but worse in terms of recall compared to the PCA method. 

\begin{table}[h!]
    \centering
    \begin{tabular}{lcccccc}
        \hline \hline
        \textbf{Algorithm} & \textbf{Accuracy} & \textbf{F1 Score} & \textbf{Recall} & \textbf{Precision}   & \textbf{AUC}  & \textbf{CT} \\
        \hline \hline
        Convex Hull, Naive + PCA     & \textbf{0.9485} & 0.1452 & 0.1577 & 0.1460  & 0.6109 & 20.1798 \\
        Convex Hull, Naive + t-SNE   & 0.9462 & 0.1661 & 0.2022 & 0.1461   & 0.5940 & 17.9756 \\
        Convex Hull, Elbow + PCA     & 0.9507 & 0.1134 & 0.1194 & 0.1470   & 0.6144 & 3.4130 \\ 
        Convex Hull, Elbow + t-SNE   & 0.9462 & 0.1657 & 0.2015 & 0.1457   & 0.5989 & 19.6112 \\
        Convex Hull, Optimal + PCA   & \textbf{0.9485} & 0.1452 & 0.1577 & 0.1460  & 0.6109 & 19.6880  \\
        Convex Hull, Optimal + t-SNE & 0.9464 & \textbf{0.1698} & 0.2067 & \textbf{0.1491} & 0.5062 & 17.7123 \\ \hline
        Isolation Forest            & 0.8856 & 0.0899 & 0.4992 & 0.0579  & 0.8871 & 1.0643 \\
        Local Outlier Factor        & 0.8886 & 0.1097 & 0.3523 & 0.0726  & 0.7841 & 0.5543 \\
        One-Class SVM               & 0.7750 & 0.0569 & 0.4106 & 0.0349  & 0.7932 & 2.0343 \\
        Mean Shift                  & 0.0760 & 0.0484 & \textbf{0.9995} & 0.0261  & 0.4985 &  2.8032 \\
        Gaussian Mixture Models     & 0.4577 & 0.0449 & 0.5820 & 0.0249  & \textbf{0.8992} & 0.5632  \\
        K-means                     & 0.4572 & 0.0459 & 0.6448 & 0.0253  & 0.4778 &  \textbf{0.1930} \\
        DBSCAN                      & 0.0344 & 0.0478 & 0.9927 & 0.0259  & 0.5070 &  0.2636 \\
        \hline \hline
    \end{tabular}
    \caption{Performance of the models with comparison to baseline algorithms for the four datasets detected as Convex Hull friendly. The best outcome for each metric is highlighted in bold.}
    \label{table:p2}
\end{table}

\subsection{Sensitivity analysis}
We computed the one-dimensional sensitivity analysis \cite{peter2010numerical} of each of the parameters. Table \ref{table:sensitivity} outlines the results of the analysis, presenting mean change obtained from a linear regression \cite{seber2012linear} model fitted on all the datasets ranging between 50\% and 100\% of the data with the smallest step size for each feature from the original value presented in Table \ref{table:datasets}. We obtained a coefficient of determination of \(R^2 = 0.906\), indicating the values of the linear regression well capture the sensitivity of the model's performance, in terms of F1 score, to each of the model's parameters. Peculiarly, only the number of samples (rows), dimensionality reduction dimension, and stopping condition are statistically significant with \(p < 0.05\).  

\begin{table}[h]
    \centering
    \begin{tabular}{lcc}
        \hline \hline
        \textbf{Parameter} & \textbf{Value} & \textbf{p-value}  \\
        \hline \hline
        \# of samples & \(0.032\) & \(4.2 \cdot 10^{-5}\) \\
        \# of features & \(-0.063\) & \(7.0 \cdot 10^{-3}\) \\
        Portion of anomalies & \(0.008\) & \(0.128\) \\
        CH-friendly & \(0.049\) & \(0.038\) \\
        Stopping condition - Naive & \(-0.005\) & \(0.046\) \\
        Stopping condition - elbow & \(0.001\) & \(0.033\) \\
        Dimensionality reduction - PCA & \(-0.082\) & \(0.083\) \\
        Dimensionality reduction dimension & \(-0.004\) & \(0.001\) \\
        \hline \hline
    \end{tabular}
    \caption{Sensitivity analysis of the dataset and proposed model parameters. The table provides parameter values and their p-values from a linear regression model predicting the model's average F1 score across the datasets in Table \ref{table:datasets}.}
    \label{table:sensitivity}
\end{table}

\review{In addition, in order to assess the robustness of the proposed CH anomaly detection algorithm against noisy data, we conducted a noise resilience analysis. Specifically, we introduced Gaussian noise to the datasets at levels ranging from 0\% to 5\% of the data variance (with mean equal to zero). Table \ref{table:noise} presents the averaged results across multiple runs for the optimal configuration of the CH algorithm (t-SNE + optimal stopping condition). Each metric is reported as a function of the percentage of added noise. One can notice that the \(F_1\) score shows a slight decline (from 0.1814 at 0\% noise to 0.1604 at 5\% noise), the precision remains relatively high, and recall exhibits only mild fluctuations. Accuracy remains consistently above 95\%, and the AUC stays in the range of 0.57-0.59. Computation time increases steadily with the noise level.}

\begin{table}[h!]
\centering
\begin{tabular}{lcccccc}
\hline \hline
\textbf{Metric / noise level} & \textbf{0\%} & \textbf{1\%} & \textbf{2\%} & \textbf{3\%} & \textbf{4\%} & \textbf{5\%} \\
\hline\hline
F1 Score    & 0.1814 & 0.1607 & 0.1683 & 0.1616 & 0.1667 & 0.1604 \\
Recall      & 0.2081 & 0.1695 & 0.1889 & 0.1847 & 0.1918 & 0.1814 \\
Precision   & 0.4934 & 0.5321 & 0.5084 & 0.4929 & 0.4757 & 0.4733 \\
Accuracy    & 0.9587 & 0.9628 & 0.9602 & 0.9602 & 0.9587 & 0.9596 \\
AUC         & 0.5934 & 0.5770 & 0.5850 & 0.5830 & 0.5857 & 0.5810 \\
CT (sec)    & 49.88  & 52.43  & 53.31  & 54.07  & 55.02  & 55.33 \\
\hline \hline
\end{tabular}
\caption{Noise resilience analysis of the Convex Hull (t-SNE + optimal stop condition) model. Metrics are averaged across \(n=10\) repetitions and all datasets in Table \ref{table:datasets} with noise levels ranging from 0\% to 5\%.}
\label{table:noise}
\end{table}

\subsection{Special cases}
\begin{figure}[!ht]
    \centering
    \begin{subfigure}[b]{0.32\textwidth}
        \centering
        \includegraphics[width=\linewidth]{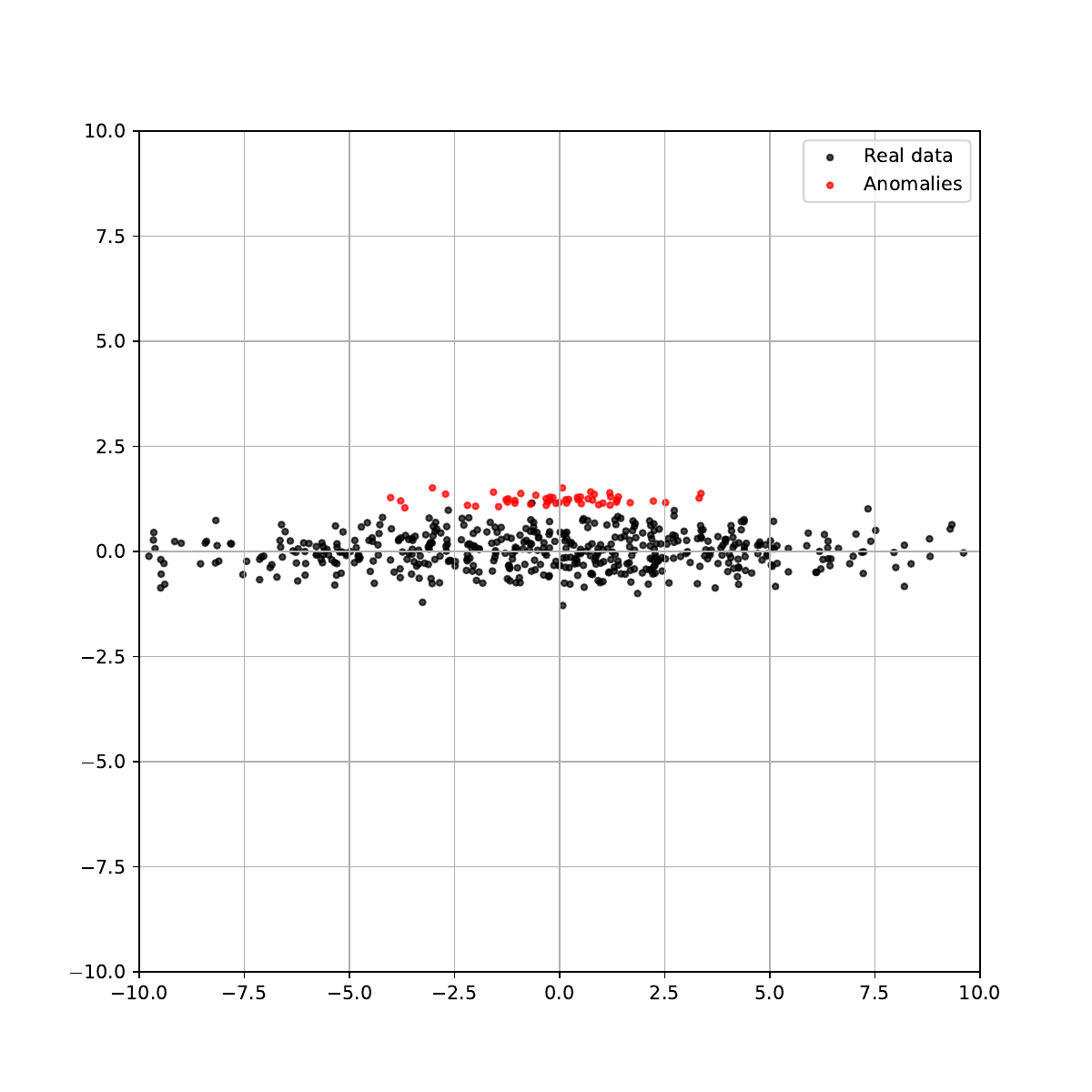}
        \caption{Unnormalized dimensions.}
        \label{subfig:s3}
    \end{subfigure}
    \begin{subfigure}[b]{0.32\textwidth}
        \centering
        \includegraphics[width=\linewidth]{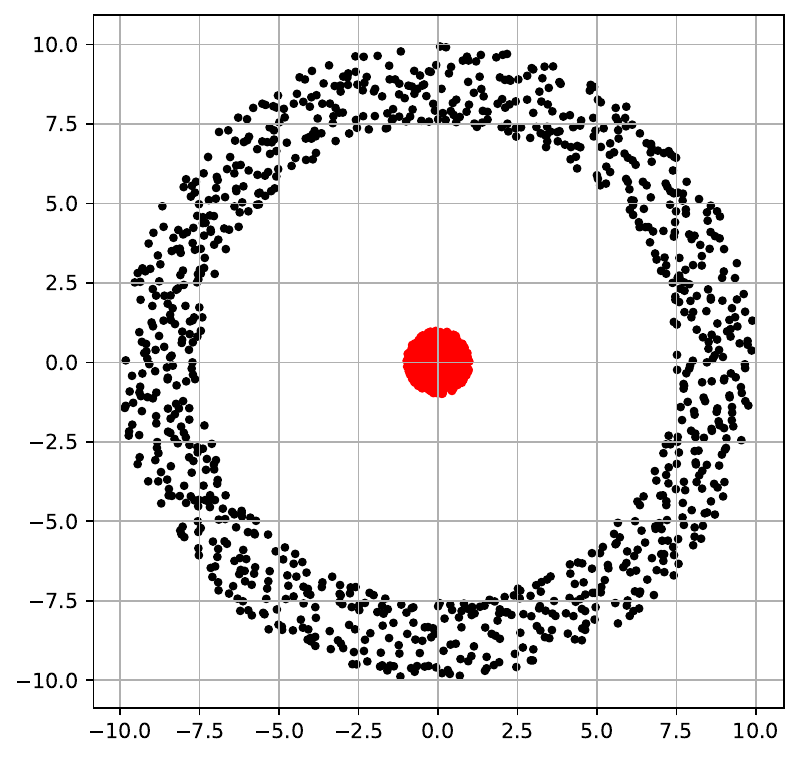}
        \caption{Torus.}
        \label{subfig:s1}
    \end{subfigure}
    \begin{subfigure}[b]{0.32\textwidth}
        \centering
        \includegraphics[width=\linewidth]{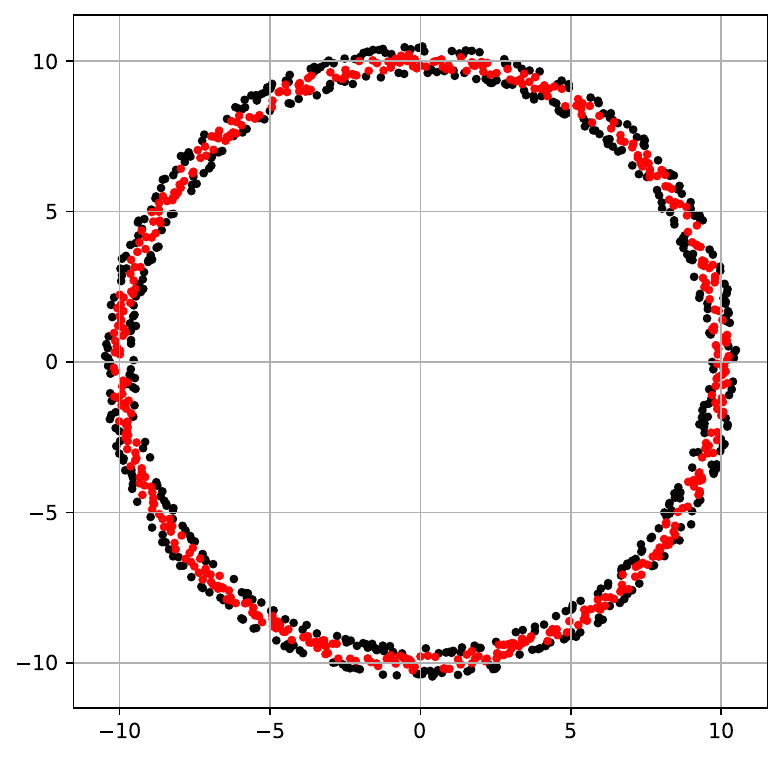}
        \caption{Circle with noise.}
        \label{subfig:s2}
    \end{subfigure}
    \caption{Special cases where the proposed Convex Hull algorithm produces poor results. The black dots indicate the in-distribution, while the red ones indicate anomalies.}
    \label{fig:subfigures}
\end{figure}

In Fig.~\ref{subfig:s3}, the convex hull provides an accurate representation of the geometric structure of the data. Consequently, the convex hull method performs effectively in this case. In contrast, Figs.~\ref{subfig:s1} and \ref{subfig:s2} illustrate cases involving a \textit{torus} and a \textit{circle} with noise, respectively. These cases are synthetically designed to highlight edge cases where the proposed Convex Hull model exhibits suboptimal performance. The primary reason for this limitation is that the majority of points in these sets lie on the convex hull's boundary. As a result, processing these boundary points demands substantial computational resources. However, it is important to note that the cases are relatively rare in most real-world datasets.

Table \ref{table:special} presents the \(F_1\) score of the Convex Hull algorithm with the optimal stopping strategy compared to the Isolation Forest algorithm for the three edge cases shown in Figs. \ref{subfig:s1}, \ref{subfig:s2}, and \ref{subfig:s3}. For the torus case the Convex Hull receives \(F_1\) score of 0 while the Isolation Forest achieves \(F_1\) score of 1. The latter two cases show less dramatic differences, with around 0.18 and 0.1 differences in the \(F_1\) score between the Isolation Forest and the Convex Hull algorithms. 

\begin{table}[h!]
    \centering
    \begin{tabular}{llc}
        \hline \hline
        \textbf{Algorithm} & \textbf{Case} & \textbf{F1 Score}  \\
        \hline \hline
        Convex Hull, Optimal & Torus & 0.0000  \\
        Convex Hull, Optimal & Circle with noise & 0.2393 \\
        Convex Hull, Optimal & Unnormalized dimensions & 0.0823 \\ \hline
        Isolation Forest & Torus & 1.0000 \\
        Isolation Forest & Circle with noise & 0.4170 \\
        Isolation Forest & Unnormalized dimensions & 0.1874  \\
        \hline \hline
        \label{table:special}
    \end{tabular}
    \caption{\(F_1\) score of the Convex Hull algorithm with the optimal stopping strategy compared to the Isolation Forest algorithm for the three edge cases shown in Figs. \ref{subfig:s1}, \ref{subfig:s2}, and \ref{subfig:s3}.}
    \label{table:special}
\end{table}

\section{Conclusions}
\label{sec:conclusions}
In this study, we introduce a volume-based Convex Hull method for anomaly detection. The approach effectively identifies anomalies by leveraging the geometric properties of data distribution, prioritizing compact and dense subsets while excluding anomalies. This dual-objective strategy ensures that anomalies, which disproportionately increase the convex hull's volume, are systematically identified. However, balancing the competing objectives of minimizing volume and maximizing data inclusion is a non-trivial challenge that requires careful calibration of parameters and stopping conditions. 

The computational complexity of the proposed method is significant, with a worst-case time complexity of $\mathcal{O}\left(N^4\right)$, where $N$ denotes the number of data points in the dataset. \review{This is between one and two order(s) of magnitude compared to the compared algorithms (see Table \ref{table:appendix} in the appendix for full time complexity information).} This high computational cost primarily stems from the iterative nature of the algorithm and the repeated calculation of convex hulls. However, the integration of dimensionality reduction techniques, such as Principal Component Analysis (PCA) and t-Distributed Stochastic Neighbor Embedding (t-SNE), as demonstrated in the experimental section, plays a critical role in alleviating these computational burdens. These techniques enable the method to achieve performance that is often comparable to, and occasionally surpasses, that of current state-of-the-art anomaly detection algorithms. Nonetheless, it is important to acknowledge the trade-offs associated with dimensionality reduction, particularly the potential distortion of the original data structure, which may affect detection accuracy.

Furthermore, the choice of stopping criteria is pivotal to the method’s overall effectiveness. The algorithm's performance varies under different stopping strategies, including naive, elbow-based, and theoretically optimal approaches. While the naive strategy is straightforward to implement, its dependence on fixed thresholds may lead to either premature termination or excessive computation. The elbow method offers a more balanced alternative by dynamically responding to changes in the convex hull's volume. Although the optimal stopping rule is impractical for real-world deployment, it provides a valuable benchmark for assessing the algorithm’s theoretical performance ceiling.

The proposed algorithm considers the geometric properties of the set, particularly the importance of distances and angles. It remains invariant under geometric transformations such as translation, rotation, and reflection. In addition, it is invariant under isotropic scaling. However, transformations that do not preserve angles, such as shearing, can alter the algorithm's outcome. It is important to note that in applications where angle preservation is critical, this behavior is acceptable.

Interestingly, the proposed algorithm produces poor results for \textit{inside} out of distribution cases \cite{lazebnik2024introducing}, as indicated by the \textit{torus}-shaped data with several anomaly data points in the center of the torus, unlike other algorithms such as the Isolation Forest which takes into account only the topology, not the distance and the angles. as the latter portions the feature space locally rather than globally like the proposed convex hull algorithm.  

This study is not without its limitations. First, the selection of stopping conditions remains a critical yet inherently subjective component of the proposed method. Although adaptive techniques such as the elbow method introduce a degree of flexibility, there is currently no universally accepted criterion for determining optimal stopping points across diverse datasets. This represents a compelling direction for future research. Second, the study does not extensively examine the effects of noise or overlapping anomalies, both of which may further challenge the robustness of the approach. Lastly, the experimental evaluation is primarily confined to small- and medium-scale real-world datasets. As a result, the scalability and generalizability of the method to large, complex datasets remain largely unexplored. This limitation is largely attributable to the scarcity of high-quality, labeled anomalies in such datasets. Addressing this gap should be a priority for future investigations. become available. 

Taken jointly, the findings of this study highlight the potential of the convex hull-based approach as a robust and interpretable method for anomaly detection across diverse datasets. By leveraging geometric principles, the method effectively identifies anomalies while offering insights into the underlying structure of the data. Despite its computational intensity and sensitivity to certain data distributions, the proposed framework demonstrates a promising balance between accuracy and interpretability, particularly when combined with dimensionality reduction techniques. 

\section*{Declarations}
\subsection*{Funding}
This study received no external funding.

\subsection*{Conflicts of Interest/Competing Interests}
The authors declare no conflict of interest.

\subsection*{Data and code availability}
The data is freely available in the following GitHub repository: \url{https://github.com/asaelbarilan/Anomaly_Detection_Using_Convex_Hull}.

\subsection*{Author contribution}
Uri Itai: Conceptualization, Formal analysis, Writing - Original Draft. \\
Asael Bar Ilan: Software, Formal analysis, Data Curation, Writing - Review \& Editing. \\
Teddy Lazebnik: Conceptualization, Methodology, Validation, Formal analysis, Software, Investigation, Data Curation, Writing - Original Draft, Writing - Review \& Editing, Supervision, Visualization, Project administration. 

\bibliography{biblio}
\bibliographystyle{unsrt}

\section*{Appendix}
\subsection*{Baseline algorithm's hyper parameters}
The baseline algorithms were configured with the following hyperparameters. For the Isolation Forest, we set \texttt{contamination=0.1} and \texttt{random\_state=42}. The One-Class SVM was used with \texttt{nu=0.1}, \texttt{kernel="rbf"}, and \texttt{gamma=0.1}. The Gaussian Mixture Models employed \texttt{n\_components=2} and \texttt{covariance\_type="full"}. K-means was run with \texttt{n\_clusters=2} and \texttt{random\_state=42}. The Local Outlier Factor used \texttt{n\_neighbors=20} and \texttt{contamination=0.1}. DBSCAN was applied with \texttt{eps=0.5} and \texttt{min\_samples=5}, while Mean Shift relied on its default parameters. 

The PCA and TSNE algorithms were constructed via \texttt{\_initialize\_reducer()}, which returns either \texttt{PCA(n\_components=2)} or \texttt{TSNE(n\_components=2)} with all other TSNE parameters left at scikit-learn defaults (e.g., \texttt{perplexity=30}, \texttt{learning\_rate="auto"}, and no fixed \texttt{random\_state}).

\subsection*{Baseline algorithm's time and memory requirements}
Table \ref{table:appendix} summarizes the computational requirements of the baseline anomaly detection algorithms in terms of average- and worst- case time complexity, as well as memory consumption. Here, \(n\) denotes the number of samples in the dataset and \(f\) denotes the number of features. 

\begin{table}[ht]
\centering
\begin{tabular}{lllc}
\hline \hline
\textbf{Algorithm} & \textbf{Time Complexity (Worst / Average)} & \textbf{Memory} & \textbf{Source} \\
\hline \hline
Isolation Forest & \(O(nlog(n)) / O(nlog(n))\) & \(O(log(f))\) & \cite{TOKOVAROV2022433} \\
Local Outlier Factor & \(O(n^2) / O(nlog(n))\) & \(O(nf)\) & \cite{bdcc5010001} \\
One-Class SVM & \(O(n^3) / O(n^3)\) & \(O(nf)\) & \cite{KANG2019772} \\
Mean Shift & \(O(n^3) / O(n^2 log(n) )\) & \(O(nf)\) & \cite{CUI2011265} \\
Gaussian Mixture Models & \(O(nf^2 + f^3) / O(nf^2 + f^3)\) & \(O(nf + f^2)\) & \cite{9625003} \\
K-means & \(O(n^2 f) / O(nf)\) & \(O(nf)\) & \cite{7065640} \\
DBSCAN &  \(O(n^2) / O(log(n))\) & \(O(nf)\) & \cite{CHENG2024120731} \\
\hline \hline
\end{tabular}
\caption{Time and memory complexity of the used anomaly detection algorithms, where \(n\) is the dataset's sample size, \(f\) is the dataset's feature size.}
\label{table:appendix}
\end{table}

\end{document}